\documentclass[11pt]{article}
\usepackage{coling2016}
\usepackage{times}
\usepackage{url}
\usepackage{latexsym}

\usepackage{amsmath,bm}
\usepackage{amsfonts}
\usepackage{mathrsfs}
\usepackage{amsbsy,amsthm,amssymb}

\usepackage{picinpar,graphicx}
\usepackage{xcolor}
\usepackage{multirow}
\usepackage{paralist}
\usepackage{subfigure}
\usepackage{arydshln}
\usepackage{CJK}

\usepackage{fancyhdr}

\title{Improved Relation Classification by Deep Recurrent Neural Networks\\
 with Data Augmentation}

\author{Yan Xu,$^{1,}$\thanks{\ \ Equal contribution.\quad $^\dag$Corresponding authors.
\quad $^\ddag$Yan Xu is currently a research scientist at Inveno Co., Ltd.\quad\hspace{1cm}{\color{white}{.}} 
{\color{white}{.}}\hspace{.4cm} $^1$Code released on \texttt{https://sites.google.com/site/drnnre/}}\ \ $^{\!,\ddag}$\ \  Ran Jia,$^{1,*}$ Lili Mou,$^1$ Ge Li,$^{1,\dag}$ Yunchuan Chen,$^2$ Yangyang Lu,$^1$ Zhi Jin$^{1,\dag}$\\
$^{1}$Key Laboratory of High Confidence Software Technologies (Peking University),\\ Ministry of Education, China;\quad Institute of Software, Peking University\\
{\tt \{xuyan14,lige,luyy11,zhijin\}@sei.pku.edu.cn}\\
{\tt\{jiaran1994,doublepower.mou\}gmail.com}\\
$^{2}$University of Chinese Academy of Sciences\ \ { \texttt{chenyunchuan11@mails.ucas.ac.cn}}\\ 
}
\date{}

\begin{document}
\maketitle
\begin{abstract}
  Nowadays, neural networks play an important role in the task of relation classification.
By designing different neural architectures, researchers have improved the performance to a large extent in comparison with traditional methods. However, existing neural networks for relation classification are usually of shallow architectures (e.g., one-layer convolutional neural networks or recurrent networks). They may fail to explore the potential representation space in different abstraction levels. In this paper, we propose deep recurrent neural networks (DRNNs) for relation classification to tackle this challenge. Further, we propose a data augmentation method by leveraging the directionality of relations. We evaluated our DRNNs on the SemEval-2010 Task~8, and achieve an $F_1$-score of 86.1\%, outperforming previous state-of-the-art recorded results.$^1$
\end{abstract}

\section{Introduction}
\blfootnote{
    %
    %
    %
    %
    %
    %
    \hspace{-0.65cm}  
    This work is licenced under a Creative Commons 
    Attribution 4.0 International License.
    License details:
    \url{http://creativecommons.org/licenses/by/4.0/}
}
Classifying relations between two entities in a given context is an important task in natural language processing (NLP). Take the following sentence as an example: ``Jewelry and other smaller [valuables]$_{e_1}$ were locked in a [safe]$_{e_2}$ or a closet with a deadbolt.'' The marked entities \textit{valuables} and \textit{safe} are of relation {\tt Content-Container}$({e_1}, {e_2})$. Relation classification plays a key role in various NLP applications, and has become a hot research topic in recent years.

Nowadays, neural network-based approaches have made significant improvement in relation classification, compared with traditional methods based on either human-designed features \cite{MaxEntRE,2010SVM} or kernels \cite{SpdKernel,EmbedTreeK}. For example, \newcite{CNN} and \newcite{CNN-NG} utilize convolutional neural networks (CNNs) for relation classification. \newcite{SDP-LSTM} apply long short term memory (LSTM)-based recurrent neural networks (RNNs) along the shortest dependency path. \newcite{EnsembleNN} build ensembles of gated recurrent unit (GRU)-based RNNs and CNNs.

\begin{figure}[!t]
\centering
\includegraphics[width=\textwidth]{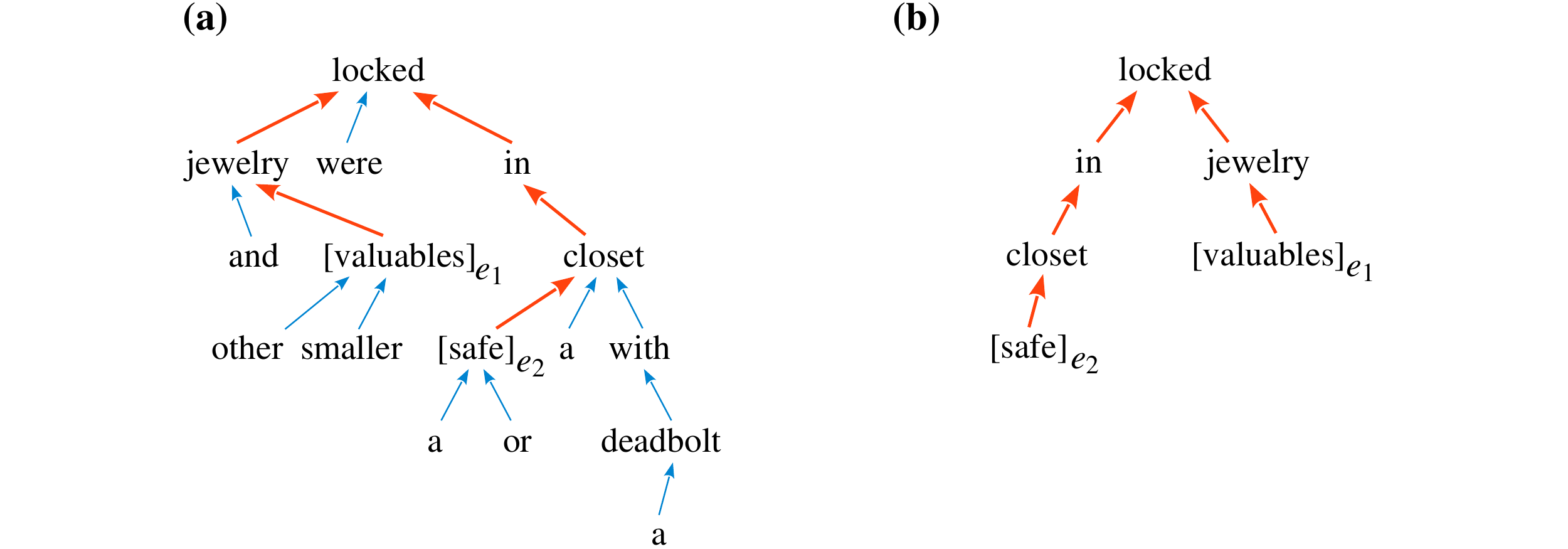}
\caption{(a) The dependency parse tree corresponding to the sentence ``Jewelry and other smaller [valuables]$_{e_1}$ were locked in a [safe]$_{e_2}$ or a closet with a deadbolt.''
Red arrows indicate the shortest dependency path between $e_1$ and $e_2$. (b) The augmented data sample.}\label{fig:example}
\end{figure}

We have noticed that these neural models are typically designed in shallow architectures, e.g., one layer of CNN or RNN, whereas evidence in the deep learning community suggests that deep architectures are more capable of information integration and abstraction \cite{SpeechDRNN,TrainDRNN,OpinionDRNN}. A natural question is then whether such deep architectures are beneficial to the relation classification task.

In this paper, we propose the deep recurrent neural networks (DRNNs) to classify relations. The deep RNNs can explore the representation space in different levels of abstraction and granularity. By visualizing how RNN units are related to the ultimate classification, we demonstrate that different layers indeed learn different representations: low-level layers enable sufficient information mix, while high-level layers are more capable of precisely locating the information relevant to the target relation between two entities. Following our previous work \cite{SDP-LSTM}, we leverage the shortest dependency path~(SDP, Figure~\ref{fig:example}) as the backbone of our RNNs.


We further observe that the relationship between two entities are directed. Two sub-paths, separated by entities' common ancestor, can be mapped to {\tt subject-predicate} and {\tt object-predicate} components of a relation. By changing the order of these two sub-paths, we obtain a new data sample with the inversed relationship (Figure~\ref{fig:example}b). Such data augmentation technique can provide additional data samples without using external data resources.


We evaluated our proposed method on the SemEval-2010 relation classification task. Even if we do not apply data augmentation, the DRNNs model has achieved a high performance of 84.2\% $F_1$-score with a depth of 3, but the performance decreases when the depth is too large. This is because the deep RNN is a large model, which necessitates more data samples for training. Applying data augmentation can alleviate the problem of data sparseness and sustain a deeper RNN to improve the performance to 86.1\%.
The results show that both our deep networks and the data augmentation strategy have contributed to the relation classification task, and that they are coupled well together for further performance improvement.

The rest of this paper is organized as follows. Section~\ref{sRelatedwork} reviews related work; Section~\ref{sModel} describes our DRNNs model in detail. Section \ref{sExperiment} presents in-depth experimental results. Finally, we have conclusion in Section \ref{sConclusion}.

\section{Related Work}\label{sRelatedwork}

Traditional methods for relation classification mainly fall into two groups: feature-based or kernel-based. The former approaches extract different types of features and feed them into a classifier, e.g., a maximum entropy model~\cite{MaxEntRE}. Various features, including lexical, syntactic, as well as semantic ones, are shown to be useful to relation classification~\cite{2010SVM}. By contrast, kernel-based methods do not have explicit feature representations, but require predefined similarity measure of two data samples. \newcite{SpdKernel} design a kernel along the shortest dependency path (SDP) between two entities by observing that the relation strongly relies on SDPs.
\newcite {EmbedTreeK} combine structural information and semantic information in a tree kernel.

Neural networks have now become a prevailing technique in this task.
\newcite{RAE} design a recursive neural network along the constituency parse tree.
\newcite{CustomRNN}, also on the basis of recursive networks, emphasize more on important phrases;
\newcite{chainRNN} restrict recursive networks to SDP.
In our previous study~\cite{SDP-LSTM}, we introduce SDP-based recurrent neural network to classify relations.

\newcite{CNN}, on the other hand, apply CNNs to relation classification.
Along this line, \newcite{RankCNN} replace the common softmax loss function with a ranking loss in their CNN model.
\newcite{CNN-NG} design a negative sampling method for SDP-based CNNs.

Besides,  representative hybrid models of CNNs and recursive/recurrent networks include \newcite{DepNN} and \newcite{EnsembleNN}.

\section{The Proposed Methodology}\label{sModel}

In this section, we describe our methodology in detail. Subsection~\ref{ssOverview} provides an overall picture of our DRNNs model. Subsections~\ref{ssRNN} and \ref{ssDeepRNN} describe deep recurrent neural networks. The proposed data augmentation technique is introduced in Subsection~\ref{ssDataAug}. Finally, we present our training objective in Subsection~\ref{ssObjective}.

\subsection{Overview}\label{ssOverview}

\begin{figure*}
\centering
\bigskip
\includegraphics[width=.9\textwidth]{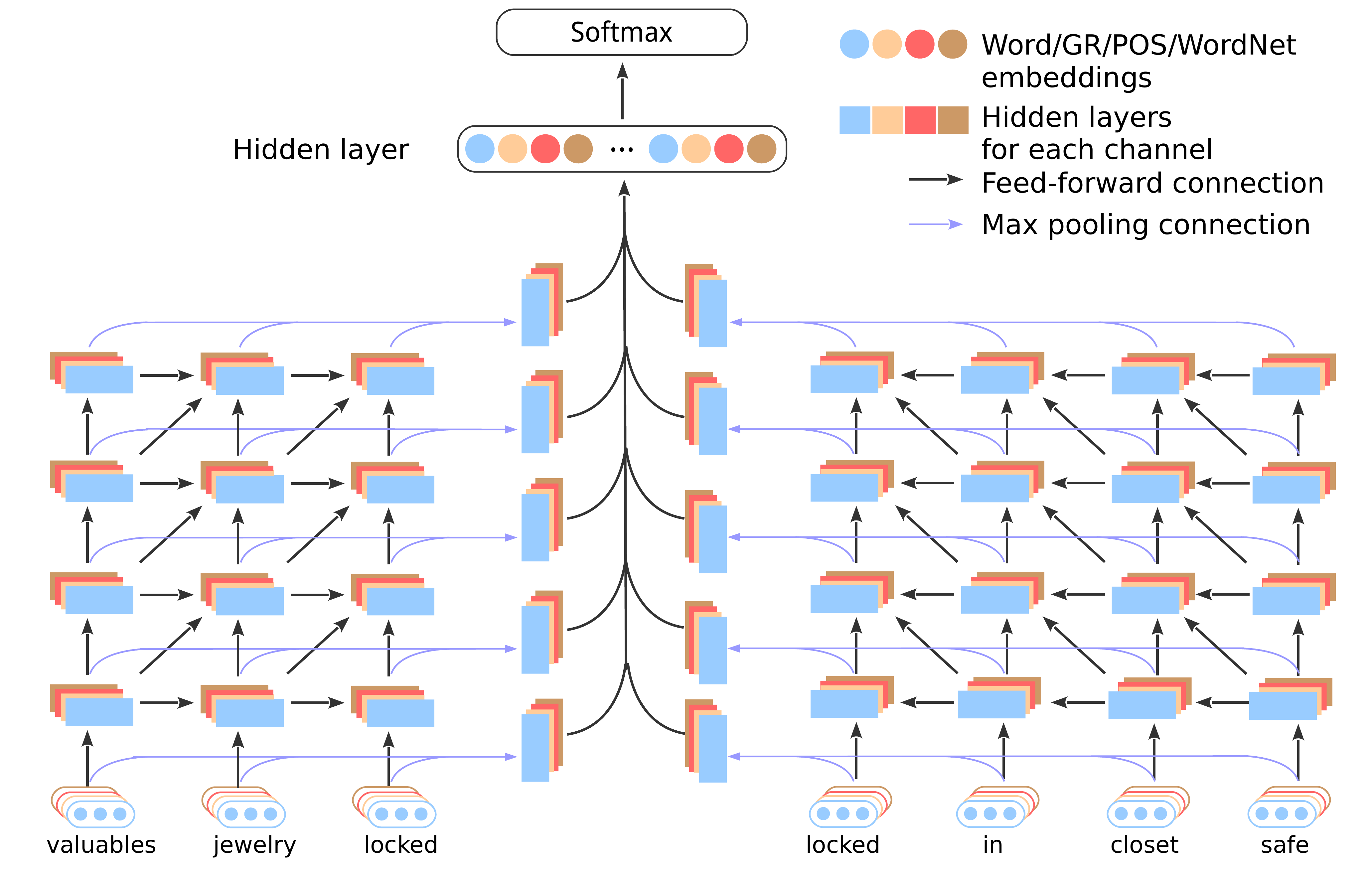}
\caption{The overall architecture of DRNNs.
Two recurrent neural networks pick up information along the shortest dependency path, separated by its common ancestor. We use four information channels, namely words, part-of-speech tags, grammatical relations (GR), and WordNet hypernyms.}\label{fArchitecture}
\end{figure*}

Figure~\ref{fArchitecture} depicts the overall architecture of the DRNNs model.
Given a sentence and its dependency parse tree,\footnote{
Parsed by the Stanford parser \cite{TypeDep}.} we follow our previous work \cite{SDP-LSTM} and build DRNNs on the shortest dependency path (SDP), which serves as a backbone. In particular, an RNN picks up information along each sub-path, separated by the common ancestor of marked entities. Also, we take advantage of four information channels, namely, word embeddings, POS embeddings, grammatical relation embeddings, and WordNet embeddings.

Different from \newcite{SDP-LSTM}, we design deep RNNs with up to four hidden layers so as to capture information in different levels of abstraction. For each RNN layer, max pooling gathers information from different recurrent nodes. Notice that the four channels (with eight sub-paths) are processed in a similar way. Then all pooling layers are concatenated and fed into a hidden layer for information integration. Finally, we have a softmax output layer for classification.

\subsection{Recurrent Neural Networks on Shortest Dependency Path}\label{ssRNN}
In this subsection, we introduce a single layer of RNN based on SDP, serving as a building block of our deep architecture.

Compared with a raw word sequence or a whole parse tree, the shortest dependency path (SDP) between two entities has two main advantages. First, it reduces irrelevant information; second, grammatical relations between words focus on the action and agents in a sentence and are naturally suitable for relation classification. Existing studies  have demonstrated the effectiveness of SDP \cite{chainRNN,DepNN,SDP-LSTM,CNN-NG}; details are not repeated here.

Focused on the SDP, an RNN keeps a hidden state vector $\bm h$, changing with the input word at each step accordingly.
Concretely, the hidden state $\bm h_t$, for the $t$-th word in the sub-path, depends on its previous state $\bm h_{t-1}$ and the current word's embedding $\bm x_t$. For the simplicity and without loss of generality, we use vanilla recurrent networks with perceptron-like interaction, that is, the input is linearly transformed by a weight matrix and non-linearly squashed by an activation function, i.e.,
\begin{equation}
\bm h_t=f(W_\text{in}\bm x_t+W_\text{rec}\bm h_{t-1}+\bm b_h)\label{eqn:shallow}
\end{equation}
where $W_\text{in}$ and $W_\text{rec}$ are weight matrices for the input and recurrent connections, respectively.
$\bm b_h$ is a bias term, and $f$ is a non-linear activation function ($\operatorname{ReLU}$ in our experiment).

\subsection{Deep Recurrent Neural Networks}\label{ssDeepRNN}

Although an RNN, as described above, is suitable for picking information along a sequence (a subpath in our task) by its iterative nature, the machine learning community suggests that deep architectures may be more capable of information integration, and can capture different levels of abstraction.

A single-layer RNN can be viewed that it is deep along \textit{time steps}. When unfolded, however, the RNN has only one hidden layer to capture the current input, as well as to retain the information in its previous step. In this sense, single-layer RNNs are actually shallow in information processing \cite{TrainDRNN,OpinionDRNN}.

In the relation classification task, words along SDPs provide information from different perspectives. On the one hand, the marked entities themselves are informative. On the other hand, the entities' common ancestor (typically verbs) tells how the two entities are related to each other. Such heterogeneous information might necessitate more complex machinery than a single RNN layer.

Following such intuition, we investigate deep RNNs by stacking multiple hidden layers on the top of one another, that is, every layer treats its previous layer as input, and computes its activation similar to Equation~\ref{eqn:shallow}.
Formally, we have

\begin{equation}
\bm h_t^{(i)}=f(W_{\text{in}}^{(i-1)}\bm h_t^{(i-1)} + W_{\text{rec}}^{(i)}\bm h_{t-1}^{(i)}
       + W_\text{cross}^{(i-1)}\bm h_{t-1}^{(i-1)} + \bm b^{(i)})\label{eqn:deep}
\end{equation}
where the subscripts refer to time steps, and superscripts indicate the layer number.
To enhance information propagation, we add a ``cross'' connection for hidden layers ($i\ge 2$) from the lower layer in the previous time step, given by $W_\text{cross}^{(i-1)}\bm h_{t-1}^{(i-1)}$ in Equation~\ref{eqn:deep}. (See also $\nearrow$ and  $\nwarrow$ arrows in Figure~\ref{fArchitecture}).



\bigskip
\subsection{Data Augmentation}\label{ssDataAug}
Neural networks, especially deep ones, are likely to be prone to overfitting. The SemEval-2010 relation classification dataset, we use, comprises only several thousand samples, which may not fully sustain the training of deep RNNs.

To mitigate this problem, we propose a data augmentation technique for relation classification by making use of the directionality of relationships.

The two sub-paths
\begin{compactitem}
\item[] $[$valuables$]_{e_1}$ $\rightarrow$ jewelry $\rightarrow$  locked
\item[] locked $\leftarrow$ in $\leftarrow$ closet $\leftarrow$ $[$safe$]_{e_2}$
\end{compactitem}
in Figure~\ref{fig:example}, for example, can be mapped to the {\tt subject-predicate} and
{\tt object-} {\tt predicate} components in the relation {\tt Content-}{\tt Container}$(e_1,e_2)$. If we change the order of these two sub-paths, we obtain
\begin{compactitem}
\item[] $[$safe$]_{e_1}$ $\rightarrow$ closet $\rightarrow$ in $\rightarrow$ locked
\item[] locked $\leftarrow$ jewelry $\leftarrow$ $[$valuables$]_{e_2}$
\end{compactitem}
Then the relationship becomes {\tt Container-}{\tt Content}$(e_1,e_2)$, which is exactly the inverse of {\tt Content-}{\tt Container}$(e_1,e_2)$. In this way, we can augment the dataset without using additional resources.

\subsection{Training Objective}\label{ssObjective}

For each recurrent layer and embedding layer (over each sub-path for each channel), we apply a max pooling layer to gather information. In total, we have 40 pools, which are concatenated and fed to a hidden layer for information integration.

Finally, a softmax layer outputs the estimated probability that two sub-paths ($s^\text{left}$ and $s^\text{right}$) are of relation $r$. For a single data sample $i$, we apply the standard cross-entropy loss, denoted as $J(s_i^\text{left}, s_i^\text{right}, r_i)$. With the data augmentation technique, our overall training objective is
\begin{align}\nonumber
J=&\sum_{i=1}^m J(s_i^\text{left}, s_i^\text{right}, r_i)+
 J(s_i^\text{right},s_i^\text{left}, r_i^{-1})
+\lambda\sum_{i=1}^\omega\|W_i\|_F
\end{align}
where $r^{-1}$ refers to the inverse of relation $r$. $m$ is the number of data samples in the original training set. $\omega$ is the number of weight matrices in DRNNs. $\lambda$ is a regularization coefficient, and $\|\cdot\|_F$ denotes Frobenius norm of a matrix.

For decoding (predicting the relation of an unseen sample), the data augmentation technique provides new opportunities, because we can use the probability of $r(e_1, e_2)$, $r^{-1}(e_2, e_1)$, or both. Section~\ref{ssExpDataAug} provides detailed discussion.

\section{Experiments}\label{sExperiment}

In this section, we present our experiments in detail.
Subsection~\ref{ssData} introduces the dataset;
Subsection~\ref{ssSetting} describes hyperparameter settings.
We discuss the details of data augmentation in Subsection~\ref{ssExpDataAug} and
the rationale for using RNNs in Subsection~\ref{ssRNNCNN}.
Subsection~\ref{ssResult} compares our DRNNs model with other methods in the literature.
In Subsection~\ref{ssDRNNsDepth}, we have quantitative and qualitative analysis of how the depth affects our model.

\subsection{Dataset}\label{ssData}

We evaluated our DRNNs model on the SemEval-2010 Task 8 dataset, which is an established benchmark for relation classification \cite{2010SVM}. 
The dataset contains 8000 sentences for training, and 2717 for testing. We split 800 samples out of the training set for validation.

There are 9 directed relations and an undirected default relation \verb|Other|; thus, we have 19 different labels in total. However, the \verb|Other| class is not taken into consideration when we compute the official measures.

\subsection{Hyperparameter Settings}\label{ssSetting}

This subsection presents hyperparameters of our proposed model.
We basically followed the settings in our previous work \cite{SDP-LSTM}.
Word embeddings were 200-dimensional, pretrained ourselves using {\tt word2vec} \cite{Word2vce} on the Wikipedia corpus; embeddings in other channels were 50-dimensional initialized randomly.
The hidden layers in each channel had the same number of units as their embeddings (either 200 or 50); the penultimate hidden layer was 100-dimensional. An $\ell_2$ penalty of  $10^{-5}$ was also applied as in \newcite{SDP-LSTM}, but we chose the dropout rate by validation with a granularity of 5\% for our model variants (with different depths).

We also chose the depth of DRNNs by validation from the set $\{1,2,\cdots,6\}$. The 3-layer and  4-layer DRNNs yield the highest performance with and without data augmentation, respectively. Section~\ref{ssDRNNsDepth} provides both quantitative and qualitative analysis regarding the effect of depth.

We applied mini-batched stochastic gradient descent for optimization, where gradients were computed by standard back-propagation.

\subsection{Data Augmentation Details}\label{ssExpDataAug}

\begin{table}[!t]
\centering
\bigskip
\begin{minipage}{0.55\textwidth}
\centering
\begin{tabular}{lc}
\hline
\textbf{Variant of Data augmentation\quad\quad} & \textbf{$F_1$}\\
\hline
No Augmentation & 84.16\\
Augment all relations     & 83.43\\
Augment {\tt Other} only & 83.01\\
Augment directed relations only & 86.10\\
\hline
\end{tabular}
\caption{Comparing variants of data augmentation.}\label{tab:aug}
\end{minipage}~~
\begin{minipage}{.4\textwidth}
\centering
\begin{tabular}{c|cc}
\hline
 & \multicolumn{2}{c}{Depth}\\
 \cline{2-3}
 & \hspace{.3cm}1\hspace{.3cm} &\hspace{.3cm} 2\hspace{.3cm}\\
\hline
\hspace{.3cm}CNN \hspace{.3cm}&\hspace{.3cm} 84.01 \hspace{.3cm}& \hspace{.3cm}83.78\hspace{.3cm}\\
\hspace{.3cm}RNN \hspace{.3cm}&\hspace{.3cm} 84.43 \hspace{.3cm}& \hspace{.3cm}85.04\hspace{.3cm}\\
\hline
\end{tabular}
\caption{Comparing CNNs and RNNs (also using $F_1$-score as the measurement).}\label{tab:CNN}
\end{minipage}
\end{table}

As mentioned in Section~\ref{ssData}, the SemEval-2010 Task 8 dataset contains an undirected class {\tt Other} in addition to 9 directed relations (18 classes). For data augmentation, it is natural that the inversed {\tt Other} relation is also in the {\tt Other} class itself. However, if we augment all the relations, we observe a performance degradation of 0.7\% (Table~\ref{tab:aug}). We deem the {\tt Other} class contains mainly noise, and is inimical to our model. Then we conducted another experiment where we only augmented the {\tt Other} class. The result verifies our conjecture as we obtained an even larger degradation of 1.1\% in this setting.

The pilot experiments suggest that we should take into consideration unfavorable noise when performing data augmentation. In this experiment, if we reverse the directed relations only and leave the {\tt Other} class intact, the performance is improved by a large margin of 1.9\%.
This shows that our proposed data augmentation technique does help to mitigate the problem of data sparseness, if we carefully rule out the impact of noise.

During validation and testing, we shall decode the target label of an unseen data sample (with two entities  $e_1$ and $e_2$). Through data augmentation, we are equipped with the probability of $r^{-1}(e_2, e_1)$ in addition to $r(e_1, e_2)$. In our experiment, we tried several settings and chose to use $r^{-1}(e_2, e_1)$ only, because it yields the highest the validation result. We think this is probably because the {\tt Other} class brings more noise to $r$ than $r^{-1}$, as the {\tt Other} class is not augmented (and hence asymmetric).

We would like to point out that our data augmentation method is a general technique for relation classification, which is not \textit{ad hoc} to a specific dataset; that the methodology for dealing with noise is also potentially applicable to other datasets.

\subsection{RNNs vs. CNNs}\label{ssRNNCNN}

As both RNNs and CNNs are prevailing neural models for NLP, we are curious whether deep architectures are also beneficial to CNNs. We tried a CNN with a sliding window of size 3 based on SDPs, similar to~\newcite{CNN-NG}; other settings were as our DRNNs.

The results are shown in Table~\ref{tab:CNN}. We observe that a single layer of CNN is also effective, yielding an $F_1$-score slightly worse than our RNN. But the deep architecture hurts the performance of CNNs in this task. One plausible explanation is that, when convolution is performed, the beginning and end of a sentence are typically padded with a special symbol or simply zero. However, the shortest dependency path between two entities is usually not very long ($\sim$4 on average). Hence, sentence boundaries may play a large role in convolution, which makes CNNs vulnerable.

On the contrary, RNNs can deal with sentence boundaries smoothly, and the performance continues to increase with up to 4 hidden layers. (Details are deferred to Subsection~\ref{ssDRNNsDepth}.)

\bigskip
\subsection{Overall Performance}\label{ssResult}

\begin{table*}[!t]
\bigskip
\centering
\resizebox{\textwidth}{!}{
\begin{tabular}{c|l|c}
\hline
\hline
\textbf{Model} &\centering \textbf{Features} &\textbf{$F_1$}\\
\hline
\multirow{2}{*}{SVM}                & POS, WordNet, prefixes and other morphological features, & \multirow{3}{*}{82.2}\\
\multirow{2}{*}{\footnotesize\cite{2010SVM} }               & depdency parse, Levin classes, PropBank, FanmeNet,       &     \\
                   & NomLex-Plus, Google $n$-gram, paraphrases, TextRunner    &     \\
\hline
RNN  & Word embeddings               & 74.8\\
\footnotesize\cite{RAE}    & + POS, NER, WordNet             & 77.6\\
\hline
MVRNN       & Word embeddings         & 79.1\\
\footnotesize\cite{MVRNN}     & + POS, NER, WordNet           & 82.4\\
\hline
CNN    & Word embeddings    & 69.7\\
\footnotesize\cite{CNN}  & + position embeddings, WordNet     & 82.7\\
\hline
Chain CNN &    \multirow{2}{*}{Word embeddings, POS, NER, WordNet} &\multirow{2}{*}{82.7}\\
\footnotesize\cite{chainRNN}&&\\
\hline
CR-CNN              & Word embeddings             & 82.8\\
\footnotesize\cite{RankCNN}      & + position embeddings       & 84.1\\
\hline
FCM       & Word embeddings      & 80.6\\
\footnotesize\cite{FCM}    & + dependency parsing, NER     & 83.0\\
\hline
SDP-LSTM & Word embeddings & 82.4\\
\footnotesize\cite{SDP-LSTM}        & Word + POS + GR + WordNet embeddings & 83.7\\
\hline
DepNN & Word embeddings + WordNet & 83.0\\
\footnotesize\cite{DepNN}                      & Word embeddings + NER     & 83.6\\
\hline
depLCNN & Word + WordNet + words around nominals & 83.7\\
\footnotesize\cite{CNN-NG}     & + negative sampling from NYT dataset & 85.6\\
\hline
Ensemble Methods & Word+POS+NER+WordNet embeddings, CNNs, RNNs + Stacking &  83.4 \\
\footnotesize\cite{EnsembleNN}   & Word+POS+NER+WordNet embeddings, CNNs, RNNs + Voting   & 84.1\\
\hline
\multirow{2}{*}{DRNNs}          & Word+POS+GR+WordNet embeddings w/o data augmentation                                          & 84.2\\
    & + data augmentation      &\textbf{86.1}\\
\hline
\hline
\end{tabular}
}
\caption{Comparison of previous relation classification systems.}
\label{tab:result}
\end{table*}
Table~\ref{tab:result} compares our DRNNs model with previous state-of-the-art methods.\footnote{This paper was preprinted on arXiv on 14 Jan 2016.}
The first entry in the table presents the highest performance achieved by traditional feature-based methods.
\newcite{2010SVM} feed a variety of handcrafted features to the SVM classifier and achieve an $F_1$-score of 82.2\%.

Recent performance improvements on this dataset are mostly achieved with the help of neural networks.
In an early study, \newcite{MVRNN} build a recursive network on constituency trees, but achieve a performance worse than \newcite{2010SVM}.
They extend their recursive network with matrix-vector interaction and elevate the $F_1$-score to 82.4\%.
\newcite{chainRNN} restrict the recursive network to SDP, which is slightly better than a sentence-wide network.
In our previous study \cite{SDP-LSTM}, we introduce recurrent neural networks based on SDP and improve the $F_1$-score to 83.7\%.

In the school of convolution, \newcite{CNN} construct a CNN on the word sequence; they also integrate word position embeddings, which benefit the CNN architecture.
\newcite{RankCNN} propose a similar CNN model, named CR-CNN, by replacing the common softmax cost function with a ranking-based cost function.
By diminishing the impact of the \verb|Other| class, they achieve an $F_1$-score of 84.1\%.
\newcite{CNN-NG} design an SDP-based CNN with negative sampling, improving the performance to 85.6\%.

Hybrid models of CNNs and RNNs do not appear to be very useful, achieving up to an $F_1$-score of 84.1\%~\cite{DepNN,EnsembleNN}.

\newcite{FCM} propose a Feature-rich Compositional Embedding Model (FCM), which combines unlexicalized linguistic contexts and word embeddings.
They do not use neural networks (at least in the usual sense) and achieve an $F_1$-score of 83.0\%.

Our DRNNs model, along with data augmentation, achieves an $F_1$-score of 86.1\%. Even if we do not apply data augmentation, the DRNNs model yields 84.2\% $F_1$-score, which is also the highest score achieved without special treatment to the noisy \texttt{Other} class. The above results show the effectiveness of DRNNs, especially trained with a large (augmented) dataset.


\subsection{Analysis of DRNNs' Depth}\label{ssDRNNsDepth}

\begin{figure*}[!t]
\centering
\includegraphics[width=.97\textwidth]{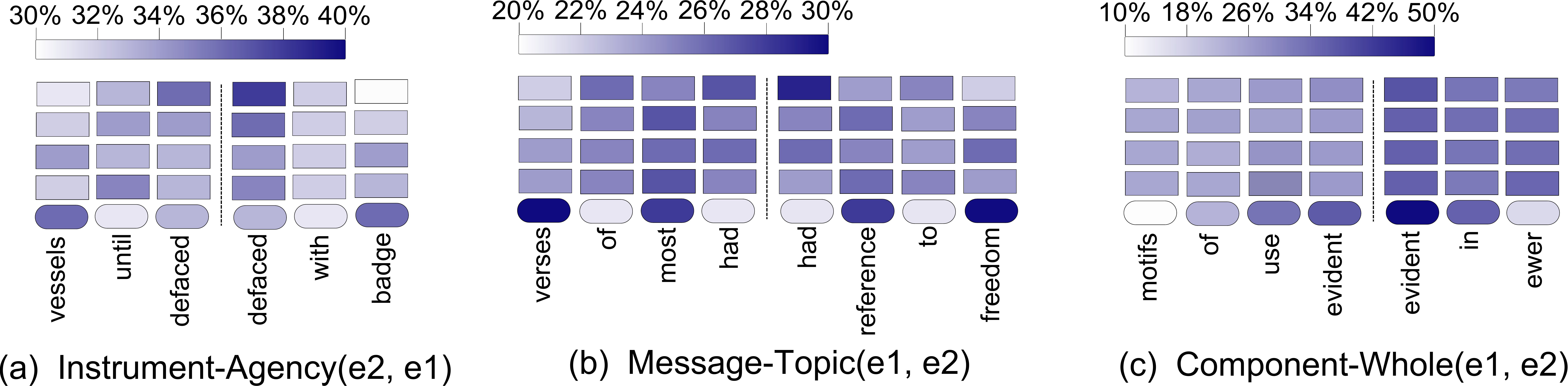}
\caption{Visualization of information propagation along multiple RNN layers.}
\label{fVis}
\end{figure*}

\begin{figwindow}[0,r,%
 \mbox{
 \centering
 \includegraphics[width=.4\textwidth]{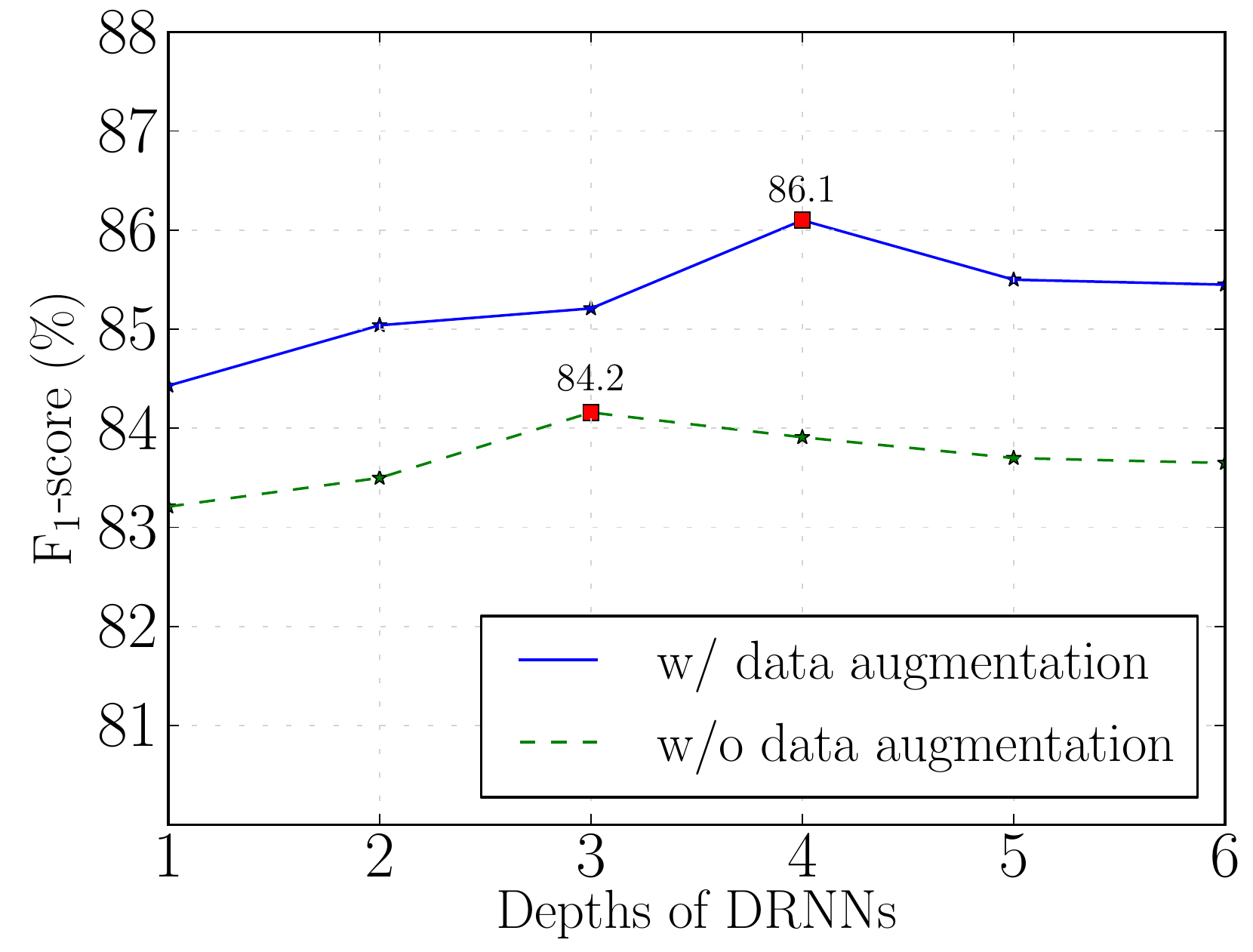}
 },
 \label{fDepEffect}
           {Analysis of the depth.$^\text{3}$
           }
 ]
In this subsection, we analyze the effect of depth in our DRNNs model.
We have tested the depth from the set $\{1, 2, \cdots, 6\}$, and plot the results in Figure~\ref{fDepEffect}. Initially, the performance increases if the depth is larger in both settings with and without augmentation. However, if we do not augment data, the performance peaks when the depth is 3. Provided with augmented training samples, the $F_1$-score continues to increase with up to 4 layers, and ends up with an $F_1$-score of 86.1\%.

\indent We next investigate how RNN units in different layers are related to the ultimate task of interest. This is accomplished by tracing back information from pooling layers. Noticing that the pooling layer takes maximum value in each dimension, we can compute how much a hidden layer's units are gathered by pooling for further processing. In this way, we are able to demonstrate the information flow in RNN hidden units. We plot three examples in Figure~\ref{fVis}. Here, rectangles refer to RNN hidden layers, unfolded along time. (Rounded rectangles are word embeddings.) The intensity of color reflects the ratio of the pooling proportion.
\end{figwindow}

\bigskip
\begin{itemize}

\item Sample 1: ``Until 1864 $[$vessels$]$$_{e_1}$ in the service of certain UK public offices defaced the Red Ensign with the $[$badge$]$$_{e_2}$ of their office'' with label \verb|Instrument-Agency|$({e_2}, {e_1})$.
    Its two sub-paths of SDP are
    \begin{compactitem}
    \item[]$\text{$[$vessels$]$}_{e_1}\rightarrow\text{until}\rightarrow\text{defaced}$
    \item[]defaced $\leftarrow$ with $\leftarrow$ $[$badge$]_{e_2}$
    \end{compactitem}
    From Figure~\ref{fVis}a, we see that entities like \textit{vessels} and \textit{badge}
    are darker than the verb phrase \textit{defaced with} on the embedding layer.
    When information is propagating horizontally and vertically, these entities are getting lighter, while the verb phrase becomes darker gradually.
    Intuitively, we think that, considering the relation \verb|Instrument-Agency|$({e_2}, {e_1})$, it is less informative with only two entities \textit{vessels} and \textit{badge}.
    When adding the semantic of verb phrase \textit{defaced with}, we are more aware of the target relation.{\color{white}\footnote{ Using vanilla RNN with a depth of 1, we obtained a slightly better accuracy in this paper than~\newcite{SDP-LSTM}.}}

\item Sample 2: ``Most of the $[$verses$]$$_{e_1}$ of the plantation songs had some reference to $[$freedom$]$$_{e_2}$'' with label \verb|Message-Topic|$({e_1}, {e_2})$.
    Its two sub-paths of SDP are
    \begin{compactitem}
    \item[] $\text{$[$verses$]$}_{e_1}\rightarrow\text{of}\rightarrow\text{most}\rightarrow\text{had}$
    \item[] had $\leftarrow$ reference $\leftarrow$ to $\leftarrow$ $[$freedom$]_{e_2}$
    \end{compactitem}
    Similar to Sample 1, we see from Figure \ref{fVis}b that the color of the ``pivot'' verb \textit{had} is getting darker vertically, and becomes the darkest in the fourth RNN layer, indicating the highest pooling portion.
    This is probably because \textit{had} links two ends of the relation, \verb|Message| and \verb|Topic|.

\item Sample 3: ``A more spare, less robust use of classical $[$motifs$]_{e_1}$ is evident in a $[$ewer$]_{e_2}$ of 1784-85'' with label \verb|Component-Whole|$({e_1}, {e_2})$.
    Its two sub-paths of SDP are
 \begin{compactitem}
    \item[]$\text{$[$motifs$]$}_{e_1}\rightarrow\text{of}\rightarrow\text{use}\rightarrow
    \text{evident}$
    \item[] evident $\leftarrow$ in $\leftarrow$ $[$ewer$]_{e_2}$
 \end{compactitem}
 Different from Figures~\ref{fVis}a and \ref{fVis}b, higher layers pay more attention to entities rather than their ancestor. In this example, \textit{motifs} and \textit{ewer} appear to be more relevant to the relation \verb|Component-|\verb|Whole| than their common ancestor \textit{evident}.
The pooling proportion of entities (\textit{motifs}, \textit{ewer}) is increasing, while other words' proportion is decreasing.

\end{itemize}

We summarize our findings as follows. (1) Pooled information usually peaks at one or a few words in the embedding layer. This makes sense because there is no information flow in this layer.  (2) Information scatters over a wider range in hidden layers, showing that the recurrent propagation does  mix information. (3) For a higher-level layer, the network pays more attention to those words that are more relevant to the relation, but whether entities or their common ancestor is more relevant is not consistent among different data samples.


\section{Conclusion}\label{sConclusion}
In this paper, we proposed deep recurrent neural networks, named DRNNs, to improve the performance of relation classification. The DRNNs model, consisting of several RNN layers, explores the representation space of different abstraction levels. By visualizing  DRNNs' units, we demonstrated that high-level layers are more capable of integrating information relevant to target relations. In addition, we have designed a data augmentation strategy by leveraging the directionality of relations.
When evaluated on the SemEval dataset, our DRNNs model results in substantial performance boost. The performance generally improves when the depth increases; with a depth of 4, our model reaches the highest $F_1$-measure of 86.1\%.

\section*{Acknowledgments}
We thank all reviewers for their constructive comments.
This research is supported by the National Basic Research Program of China (the 973 Program) under Grant No.\@ 2015CB352201, the National Natural Science Foundation of China under Grant Nos.\@ 61232015, 91318301, 61421091, and 61502014. 

\bibliographystyle{acl}
\bibliography{DeepRNN}

\end{document}